\UseRawInputEncoding
\def\year{2021}\relax
\documentclass[letterpaper]{article} 
\usepackage{aaai21}  
\usepackage{times}  
\usepackage{helvet} 
\usepackage{courier}  
\usepackage[hyphens]{url}  
\usepackage{graphicx} 
\usepackage{bm}
\usepackage{amssymb}
\usepackage{natbib}  
\usepackage{caption} 
\usepackage[switch]{lineno}
\usepackage{multirow}
\usepackage[tbtags]{amsmath}
\title{Tracking Interaction States for Multi-Turn Text-to-SQL Semantic Parsing}
\author{
    Run-Ze Wang\textsuperscript{\rm 1},
    Zhen-Hua Ling\thanks{Zhen-Hua Ling is corresponding author.} \textsuperscript{\rm 1},
    Jing-Bo Zhou\textsuperscript{\rm 2},
    Yu Hu\textsuperscript{\rm 3,4}
    \\
}
\affiliations{
    \textsuperscript{\rm 1}National Engineering Laboratory for Speech and Language Information Processing, \\
    University of Science and Technology of China\\
    \textsuperscript{\rm 2}Business Intelligence Lab, Baidu Research\\
    \textsuperscript{\rm 3}University of Science and Technology of China\\
    \textsuperscript{\rm 4}iFLYTEK Research\\

    wrz94520@mail.ustc.edu.cn, zhling@ustc.edu.cn, zhoujingbo@baidu.com,
    \ yuhu@iflytek.com

}
\begin{document}
\maketitle
\begin{abstract}
The task of multi-turn text-to-SQL semantic parsing aims to translate natural language utterances in an interaction into SQL queries in order to answer them using a database which normally contains multiple table schemas. Previous studies on this task usually utilized contextual information to enrich utterance representations and to further influence the decoding process. While they ignored to describe and track the interaction states which are determined by history SQL queries and are related with the intent of current utterance. In this paper, two kinds of interaction states are defined based on schema items and SQL keywords separately. A relational graph neural network and a non-linear layer are designed to update the representations of these two states respectively. The dynamic schema-state and SQL-state representations are then utilized to decode the SQL query corresponding to current utterance. Experimental results on the challenging CoSQL dataset demonstrate the effectiveness of our proposed method, which achieves better performance than other published methods on the task leaderboard.
\end{abstract}

\section{Instruction}
\label{intro}
\begin{figure}[t]
\centering
\includegraphics[width=0.9\columnwidth]{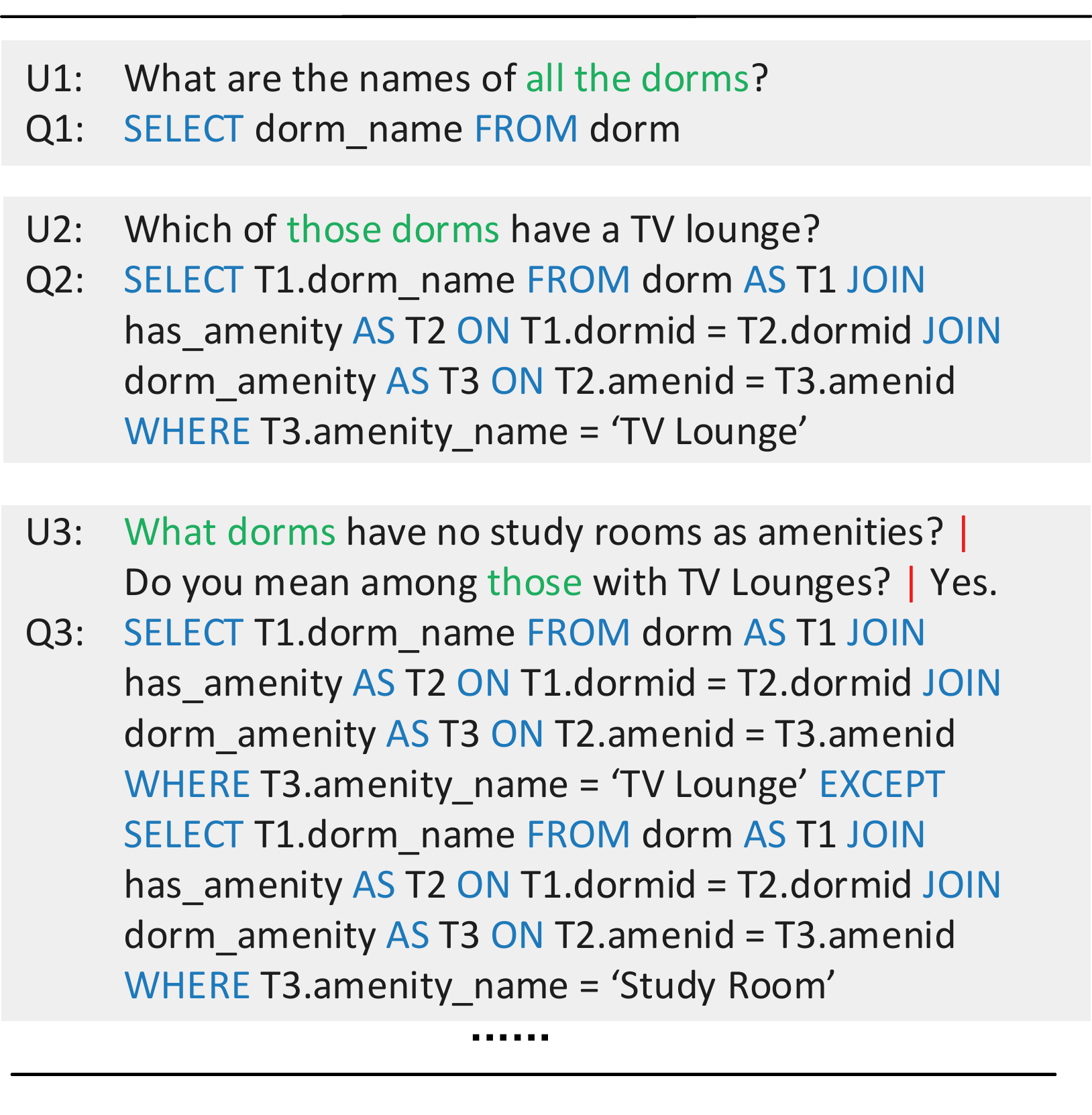} 
\caption{An interaction example from CoSQL dataset. $U_i$ is the user utterance at the $i$-th turn while $Q_i$ is its corresponding SQL query.
The blue words are SQL keywords. Because this paper only focuses on the text-to-SQL task of CoSQL, we concatenate those utterances that cannot be translated into SQL queries with their responds offered in the original CoSQL dataset using ``$|$", such as $U_3$.}
\label{example}
\end{figure}
Querying relational databases and acquiring information from it have been studied extensively when designing natural language interfaces to databases\cite{zelle1996learning,zhang2020joint}.
As such, a large body of research has focused on the task of translating natural language user utterances into SQL queries that existing database software can execute \cite{zhong2017seq2sql,xu2017sqlnet,yu2018typesql}.
While most of these works focus on precisely mapping stand-alone utterances to SQL queries \cite{krishnamurthy2017neural, dong2018coarse}, users often access information by multi-turn interactions in real-word applications.
Therefor, designing a text-to-SQL system for multi-turn scenario has attracted more and more attention recently \cite{guo2019towards, bogin2019representing, Yu19SParC, yu2018spider}.

The multi-turn text-to-SQl task is more complex and more challenging than traditional stand-alone text-to-SQL task.
The main challenge is how to utilize the contextual information in  history utterances to help system better parse current utterance.
Figure \ref{example} is an interaction example extracted from CoSQL dataset \cite{yu2019cosql}. 
We can see from this figure that the phrase ``those dorms" in $U_2$ refers to the phrase ``all the dorms" in $U_1$.
And the phrases ``what dorms" and ``those" in $U_3$ also refer to the same things as above.
Besides, each predicted SQL query has obvious overlaps with history ones.
Furthermore, each interaction corresponds to a database, called ``schema", with several tables in it.
The usage of table information in history queries also contributes to decode current query.
For example, the SQL query $Q_2$ mentions the ``dorm", ``has\_amenity" and ``dorm\_amenity" tables in the schema.
Thus, the new clause ``EXCEPT" in the SQL query $Q_3$ is more likely to select its contents from these tables.

Previous studies on multi-turn text-to-SQL usually adopted encoder-decoder architectures.
The encoder aims to represent user utterances and table schemas of databases as vectors, while the decoder is designed to generate corresponding SQL queries based on the representations given by the encoder.
\citet{suhr2018learning} designed a context-dependent model for ATIS (Airline Travel Information System) \cite{hemphill1990atis, dahl1994expanding}, which is a single-domain multi-turn text-to-SQL task.
They utilized an LSTM-based discourse state to integrate sentence-level representations of history user utterances at the encoding stage.
\citet{zhang2019editing} improved this method on SParC \cite{Yu19SParC} and CoSQL \cite{yu2019cosql} datasets.
They concatenated all history utterance tokens behind current utterance tokens and fed them into a BERT-based encoder together with schema items in order to generate user utterance and table schema representations.
At the decoding stage, both of these two methods employed a copy mechanism to make use of the tokens or segments in previous predicted queries.

This paper studies the multi-turn text-to-SQL task by comparing it to another popular NLP task, task-oriented dialogue system.
Both tasks aims at accomplish specific goals, i.e., querying databases or solving problems.
The user utterances in multi-turn text-to-SQL are similar to the user inputs in dialog systems, and SQL queries correspond to dialogue responses.
Besides, both tasks strongly rely on history information when making predictions.
An essential component in task-oriented dialogue systems is dialog state tracking (DST) \cite{DBLP:conf/eacl/Rojas-BarahonaG17, DBLP:conf/emnlp/BudzianowskiWTC18, DBLP:conf/acl/MrksicSWTY17}, which keeps track of user requests or intentions throughout a dialogue in the form of a set of slot-value pairs, i.e. dialogue states.
However, for the multi-turn text-to-SQL task, how to define, track and utilize user intentions throughout an interaction  has not yet been investigated in previous work.

Therefore, we propose a method of tracking interaction states for multi-turn text-to-SQL semantic parsing in this paper.
Considering the goal of text-to-SQL is to generate SQL queries for executing on table-based databases,
two types of interaction states (i.e., name-value pairs) are defined based on schema tables and SQL keywords respectively.
For schema-states, their names are the column names of all tables in the schema.
Their values come from SQL keywords and are extracted from the last predicted SQL query.
For example, after generating $Q_2$ in Figure~\ref{example}, the value of the schema-state ``dorm.dorm\_name" is set as \{``SELECT"\}.
After generating $Q_3$, its value is updated into a set of keywords \{``SELECT", ``=", ``EXCEPT", ``SELECT"\}.
For SQL-states, their names are all the SQL keywords. 
Their values are column names and are also determined by the last predicted SQL query.
For example, after generating $Q_2$ in Figure~\ref{example}, the value of the SQL-state ``SELECT"  is ``dorm.dorm\_name".
This value is not changed after generating $Q_3$  because the ``SELECT" keyword in the nested clause is also followed with ``dorm.dorm\_name" in $Q_3$.

In order to encode and utilize these designed interaction states,  a model of text-to-SQL parsing with interaction state tracking (IST-SQL) is designed.
In this model, each state is updated by a state updater based on last predicted query.
A relational graph neural network (RGNN) is employed to encode the schema-state representations.
A schema-column-graph is built based on the foreign keys in the database for implementing the RGNN. 
Besides, an one-layer non-linear layer is adopted to encode SQL-states.
The BERT model is used at the beginning of IST-SQL to generate the embedding vectors of current utterance  and all schema column names.
The utterance embedding is further fed into an utterance encoder to integrate the information of all history utterances.
Finally, the utterance representations, schema-state representations and SQL-state representations are fed into a decoder with copy mechanism to predict the SQL query tokens in order.

We evaluate our model on the CoSQL dataset, which is the largest and the most difficult dataset for conversational and multi-turn text-to-SQL semantic parsing.
Experimental results show that our model improves the question-matching accuracy (QM) of the previous best model \cite{zhang2019editing} from 40.8\% to 41.8\% and the interaction matching accuracy (IM) from 13.7\% to 15.2\%, respectively.

The main contributions of this paper are twofolds.
First, we propose to track interaction states for multi-turn text-to-SQL. Two types of interaction states are defined and are updated, encoded and utilized in our proposed IST-SQL model.
Second, our proposed model performs better than the advanced benchmarks on the CoSQL dataset.

\section{Related Work}
\subsection{Text-to-SQL Datasets and Methods}
Text-to-SQL generation task is one of the most important tasks in semantic parsing, which aims to map natural language utterances into formal representations, such as logical forms \cite{zelle1996learning, clarke2010driving}, lambda calculus \cite{zettlemoyer2005learning, artzi2011bootstrapping} and executable programming languages \cite{miller1996fully, yin2017syntactic}.

At the earliest time, researchers focused on the converting single utterances while their corresponding SQL queries were simple and always produced in one domain, such as GeoQuery \cite{zelle1996learning} and Overnight \cite{wang2015building}.
Recently, some large-scale datasets with  open domains database have been released to attract attentions to the unseen-domain problem \cite{bogin2019global, dong2016language,dong2018coarse,bogin2019representing, finegan2018improving}.
WikiSQL \cite{zhong2017seq2sql} is a popular open domain text-to-SQL task but the SQL queries in it are still simple, just contains ``SELECT", ``WHERE" and ``FROM" clauses.
To study the complex SQL queries, \citet{yu2018spider} released a complex cross-domain text-to-SQL dataset names Spider, which contains most of the SQL clauses.

There are also some dataset that can support the studies  on multi-turn text-to-SQL semantic parsing, such as ATIS \cite{hemphill1990atis}, SequentialQA \cite{iyyer2017search}, CoSQL \cite{yu2019cosql} and SParC \cite{yu2019sparc}.
SParC \cite{yu2019sparc} was built based on Spider by dividing its single utterance into multiple ones and writing their corresponding SQL queries by annotators.
CoSQL \cite{yu2019cosql} was released in 2019, which is the main evaluation dataset of this paper.
Its user utterances were all collected from Wizard-of-Oz (WOZ) conversations and the corresponding SQL queries were written by annotators.

Previous methods on text-to-SQL semantic parsing always adopted encoder-decoder architectures.
LSTM and BERT were popularly employed  to obtain the representations of user utterances and database items \cite{xu2017sqlnet, dong2018coarse, guo2019towards}.
Classification-based \cite{yu2018typesql, xu2017sqlnet, DBLP:conf/emnlp/YuYYZWLR18}, sequence-based \cite{zhang2019editing}, and abstract-syntax-tree-based \cite{dong2018coarse, guo2019towards, bogin2019representing} decoders have been developed for this task.
In order to deal with the multi-turn scenario, previous studies usually integrated history utterance information into the representation of current utterance and introduced copy mechanism into the decoder for utilizing previous predicted SQL queries \cite{hemphill1990atis, zhang2019editing}.
In contrast, we propose to track interaction states for multi-turn text-to-SQL and define two type of interaction states to record history information in this paper.

\subsection{Dialog State Tracking Task}
Our proposed method is inspired by the dialog state tracking (DST) component in task-oriented dialogue systems\cite{DBLP:conf/sigdial/HendersonTW14, gu2019dually, gu2019interactive}.
The goal of task-oriented dialogue systems is to help users accomplish a specific task such as hotel reservation, flight booking or travel information searching. \cite{DBLP:conf/eacl/Rojas-BarahonaG17, DBLP:conf/emnlp/BudzianowskiWTC18, DBLP:conf/acl/MrksicSWTY17}.
Dialog state tracking is to records the dialog process and the user intention at each turn \cite{DBLP:conf/acl/WangTWQY20} in task-oriented dialogue systems.
Here, dialogue states, i.e., a set of slot-value pairs, are usually pre-defined manually.
Various methods have been proposed for dialog state tracking. For example, \citet{DBLP:conf/acl/OuyangCDZHC20} used a connection model to connect current states with previous states and copy previous values. \citet{DBLP:conf/acl/HuYCHY20} designed an slot attention module and slot information sharing module for better utilizing the slot information.

In this paper, we bring the idea of  dialog state tracking to the multi-turn text-to-SQL task.
This task always studies open-domain datasets, which means that interaction states (similar to dialog states) should be grounded to different domains and can not be set universally for all interactions.
In this paper, we define two types of interaction states, schema-states and SQL-states.
Schema-states record domain-specific information and are changed with the grounded database.
SQL-states are used to record the SQL information at each turn for better understanding the interaction process.

\section{Preliminary}
\subsection{Dataset}
We evaluated our model on CoSQL \cite{yu2019cosql}, which is a large-scale conversational and multi-turn text-to-SQL semantic parsing dataset.
An interaction example in CoSQL is shown in Figure \ref{example}.
CoSQL consists of 30k+ turns plus 10k+ annotated SQL queries, obtained from a Wizard-of-Oz(WOZ) collection of 3k conversations querying 200 complex databases spanning 138 domain.
It has been separated into 2164 interactions with 140 databases for training,  292 interactions with 20 databases for development and 551 interactions with 40 databases for testing.

We also evaluate our model on SParC \cite{yu2019sparc}, which is another multi-turn text-to-SQL task.
SParC consists of 3034 interactions with 140 databases for training, 422 interactions with 20 databases for development and 841 interactions with 40 databases for testing.

In both datasets, several interactions may be grounded to the same database, but each database can only appear in one of the training, development and testing sets for cross-domain evaluation.

\subsection{Task Formulation}
Let $U$ denote a natural language user utterance and $Q$ denote its corresponding SQL query.
A multi-turn text-to-SQL task considers an interaction $I$, which consists of a sequence of $(U_t, Q_t)$ pairs.
Each interaction $I$ is grounded to a database $S$, which consists of several tables $T$ and each table $T$ consists of several column names $C$.
In our experiments, we concatenated each column name with its corresponding  table name, thus the set of table names $T$ can be omitted.

The goal of multi-turn text-to-SQL task is to generate $\{Q_t\}_{t=1}^{N(I)}$ in order given $\{U_t\}_{t=1}^{N(I)}$ and $S$ as follows,
\begin{equation}
\{\{U_t\}_{t=1}^{N(I)}, S\} \stackrel{map}{\longrightarrow} \{Q_t\}_{t=1}^{N(I)}.
\end{equation}
The function $N(*)$ in this paper stands for the element number of $*$.
In this paper, as we utilize the last predicted SQL query for extracting interaction states and for copying tokens when decoding,
the goal at the $t$-th turn of this task is to generate the $t$-th SQL query $Q_t$ given the current utterance $U_t$, all the history utterance $\{U_1, \cdots, U_{t-1}\}$, the table schema $S$ and the last predicted query $Q_{t-1}$, i.e.,
\begin{equation}
\{U_t, S, \{U_1, \cdots, U_{t-1}\}, Q_{t-1}\} \stackrel{map}{\longrightarrow} Q_t.
\end{equation}

\section{Proposed Method}
\begin{figure*}[t]
\centering
\includegraphics[width=0.85\textwidth]{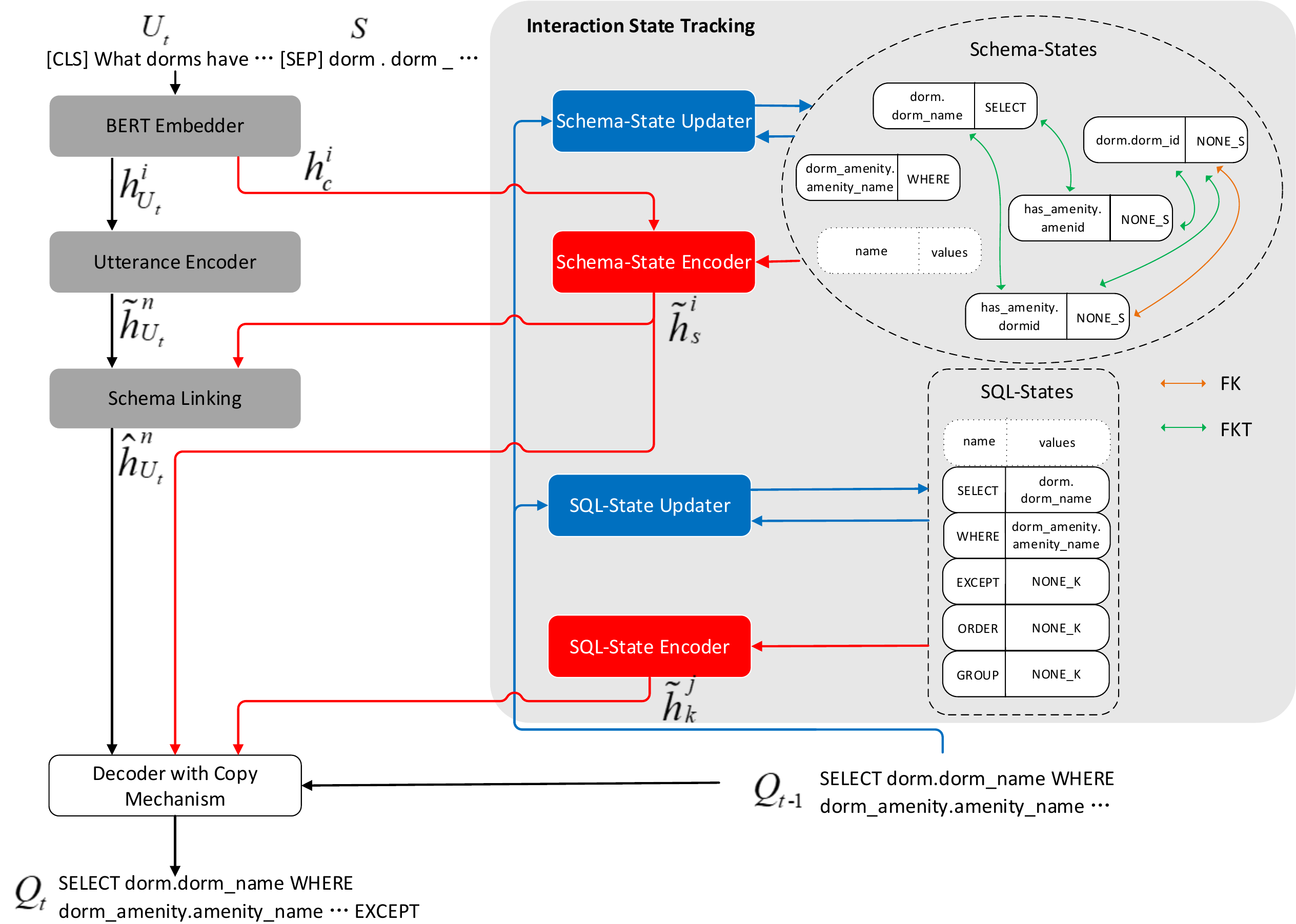}
\caption{The architecture of proposed IST-SQL model. ``FK" and ``FKT" stand for Foreign-Key and Foreign-Key-Table relations between schema-states respectively.}
\label{model}
\end{figure*}
As shown in Figure~\ref{model}, our IST-SQL model is built based on the sequence-to-sequence encoder-decoder architecture.
It consists of (1) a BERT embedder to embed current utterance tokens and schema column names into embedding vectors, (2) an utterance encoder to integrate history utterances into the representations of current utterance, (3) an interaction states tracking module with state updaters and state encoders, and (4) a decoder with copy mechanism to predicted SQL query tokens in order. 

In our model, the BERT embedder is the same as one used in \cite{zhang2019editing} and more details can be found in that paper.

\subsection{Interaction State Tracking}
\label{methods}
\paragraph{State Updaters}
As introduced in the Introduction section, two kinds of interaction states are designed in this paper.
For SQL-states, their names are SQL keywords  and their values come from the schema column names corresponding to the interaction.
For schema-states, their names are column names in the schema and their values are SQL keywords.
As shown in Figure~\ref{model}, the SQL-state updater and the schema-state updater extracted state values from the last previous predicted SQL query at each turn. 

To extract the SQL-state values at each turn, we separate the last predicted SQL query with all the column names in it.
$Q_t$ and $Q_{t-1}$ in Figure \ref{model} correspond to $Q_3$ and $Q_2$ in Figure \ref{example} respectively.
The SQL-state updater first separates $Q_{t-1}$ into a set of pieces \{``SELECT dorm.dorm", ``WHERE dorm\_amenity.amenity\_name"\}\footnote{Following previous methods, ``FROM", ``JOIN" and ``ON" clauses are removed when model predicting because they can be easily filled with some simple rules after all the other clauses are predicted.}.
If an SQL-state appears in one of these pieces, the column name appears at the end of this piece is added to the SQL-state as its value.
As shown in Figure~\ref{model}, the value of SQL-state ``SELECT" is ``dorm.dorm" and the value of ``WHERE" state is ``dorm\_amenity.amenity\_name".
It should be noticed that an SQL-state can have multiple non-repetitive values. 
The values of those SQL-states that do not appear in the last predicted SQL query are set as a fixed token ``NONE\_K" .

In a dual way, if the name of a schema-state appears in one of the SQL-state values, the schema-state updater adds the name of this SQL-state to the schema-state as its value.
In order to describe the nesting structure of SQL queries, the SQL keywords that occur more than once in the last query are all kept as schema-state values.
In Figure \ref{model}, the value of schema-state ``dorm.dorm"  is ``SELECT" while the value of ``dorm\_amenity.amenity\_name" state is ``WHERE".
The values of those schema-states  that do not appear in the last predicted SQL query are set as a fixed token ``NONE\_S" .


\paragraph{State Encoders}
The function of the schema-state encoder and the SQL-state encoder in Figure~\ref{model} is to convert the values of both states into embedding vectors at each turn.

Considering that the names of schema-states are column names in a relational table-based database and different column names are related to each other,
a relational graph neural network (RGNN) is designed to represent such relations and to propagate node information to nearby nodes.
Here, the RGNN model uses schema-states as nodes and builds edges according to their relations.
Based on foreign keys \footnote{Foreign key is a pre-defined relation between column names and stands for that these two columns has the same meanings and values but are in different tables. More details can be found in \cite{yu2018spider}.}, two kinds of relations between two schema-states are considered as follows.
\begin{itemize}
\item{\textbf{Foreign-Key (FK)}} If two column names are a foreign key pair in the database,  we build an edge with ``IN" and ``OUT" directions between them, as shown by the yellow line and arrows between schema-states in Figure \ref{model}.
\item{\textbf{Foreign-Key-Table (FKT)}} If two column names are in different tables and the two tables have one or more foreign key pairs in their columns,  we build an edge with ``IN" and ``OUT" directions between them, as shown by the green lines and arrows between schema-states in Figure \ref{model}.
\end{itemize}

The representation of each node, i.e., schema-state, is initialed with the representations of its name and values.
Because the schema-state names are all the schema column names, we directly use the column name representations $\{\bm{h}_c^i\}_{i=1}^{N(c)}$ generated by the BERT embedder in Figure \ref{model}, where $N(c)$ is the total number of schema columns.
The value embedding vectors $\{\bm{h}_k^j\}_{j=1}^{N(k)}$ are initialed randomly, where $N(k)$ is the total number of SQL keywords. 
The initial representation of each schema-state is calculated as
\begin{equation}
\bm{h}_{s}^{i} =tanh(\bm{W_1}(\bm{h}_c^i+ \sum_{j \in V_s^i} \bm{h}_k^j)),
\label{schema_update}
\end{equation}
where $V_s^i$ is the value index set of the $i$-th schema-state, $\bm{W}_1 \in \mathbb{R}^{d \times d}$ is a trainable matrix and $d$ is the hidden state dimension.

Then, $\{\bm{h}_{s}^{i}\}_{i=1}^{N(c)}$ are fed into the RGNN and the final representations are calculated as
 \begin{equation}
\bm{h}_{out}^{i} = \sum_{e \in \{FK, FKT\}} \sum_{j \in \varepsilon_{out}^i(e)} \bm{W}_{out}(\bm{z}_{out}^e * \bm{h}_{s}^{j}),
\end{equation}
\begin{equation}
\bm{h}_{in}^{i} = \sum_{e \in \{FK, FKT\}} \sum_{j \in \varepsilon_{in}^i(e)} \bm{W}_{in}(\bm{z}_{in}^e * \bm{h}_{s}^{j}),
\end{equation}
\begin{equation}
\bm{\widetilde{h}}_{s}^i =  \bm{h}_{s}^{i} + 0.5 * \bm{h}_{out}^{i} + 0.5 * \bm{h}_{in}^{i},
\end{equation}
where $\bm{W}_{out}$ and $\bm{W}_{in}$ are trainable matrices for ``IN" and ``OUT" directions, $\bm{z}_{out}^e$ and $\bm{z}_{in}^e$ stand for the ``IN" and ``OUT" embedding vectors of edge $e$, which are randomly initialized and are updated during training. $*$ stands for element-wise product. $\varepsilon_{in, out}^i(e)$ denotes the set of schema-state indices that have edge $e$ connecting with state $i$.
$\bm{\widetilde{h}}_{s}^i$ is the final schema-state representation for the $i$-th schema-state which is further used by schema linking and decoding.

Considering that the names of  SQL-states are all SQL keywords without explicit relations, we simply build a network with one non-linear layer to derive their representations as
\begin{equation}
\bm{\widetilde{h}}_{k}^{j} = tanh(\bm{W_2}(\bm{h}_k^{j}+ \sum_{i \in V_k^j} \bm{h}_s^i)),
\label{key_update}
\end{equation}
where $\bm{h}_k^{j}$ is the representation of the $j$-th SQL-state name, 
$V_k^j$ is the value index set of the $j$-th SQL-state, $\bm{W}_2 \in \mathbb{R}^{d \times d}$ is a trainable matrix.
The final SQL-state representation $\bm{\widetilde{h}}_{k}^{i}$ is further fed into the decoder as Figure \ref{model} shows.

\subsection{Utterance Encoder}
Integrating history utterance information into the representation of current utterance is important for multi-turn text-to-SQL.
Unlike previous methods that utilized history utterances at sentence-level, we construct a token-level utterance encoder based on multi-head attention mechanism to enrich utterance representations.
Let $\bm{h}_{U_t}^n$ stand for the embedding vector given by the BERT embedder for the $n$-th token in utterance $U_t$.
We collect the history information as
\begin{equation}
\bm{h}_{\{1, \cdots, t-1\} \rightarrow t}^{n} =\frac{1}{t-1}\sum_{m=1}^{t-1} \bm{h}_{m \rightarrow t}^{n},
\label{weight_ave}
\end{equation}
where
\begin{equation}
\bm{h}_{m \rightarrow t}^{n} =\sum_{k=1}^K \sum_{l=1}^{N(U_m)} \alpha^{nkl}_m \bm{h}_{U_m}^l,
\label{weight_sum}
\end{equation}
\begin{equation}
\alpha^{nkl}_m = softmax(s_m^{nkl}),\\
\end{equation}
\begin{equation}
s_m^{nkl} = (\bm{h}_{U_t}^n * \bm{a}_u^k)\cdot \bm{h}_{U_m}^l.
\label{weight_att}
\end{equation}
Here, $*$ stands for element-wise product and $\cdot$ stands for dot product, the $softmax$ function is executed over the index of $l$,
$\bm{a}_{u}^k$ is a trainable vector and $K$ is the head number.
The embedding vectors of current utterance tokens after utterance encoder are calculated as
\begin{equation}
\bm{\widetilde{h}}_{U_t}^n = BiLSTM([\bm{h}_{U_t}^n; \bm{h}_{\{1 \cdots t-1\} \rightarrow t}^{n}]),
\label{lstm}
\end{equation}
and are further fed into the schema linking module.

\subsection{Schema Linking}
Similar to previous studies \cite{guo2019towards, bogin2019representing}, we also build an schema linking module, which integrates database information into utterance representations to deal with unseen-domain problem.
Previous methods usually treated schema linking as a pre-processing step to generate some features linking utterance tokens and schema items. 
In our IST-SQL model, the schema linking module is similar to the utterance encoder.
There are two main differences.
First, $\{\bm{h}_{U_m}^l\}_{m=1}^{t-1}$ in Eq. (\ref{weight_sum}) and (\ref{weight_att}) are replaced with schema-state representation $\bm{\widetilde{h}}_{s}^i$ for calculating its relevance with utterance representation $\bm{\widetilde{h}}_{ U_t}^{n}$, and the averaging operation in Eq. (\ref{weight_ave}) is not conducted.
Second, considering $\bm{\widetilde{h}}_{U_t}^n$ and $\bm{\widetilde{h}}_{s}^i$ are embedded in different spaces, a transform vector $\bm{b}_s^k$ is applied to 
Eq. (\ref{weight_sum}) for the $k$-th head as
\begin{equation}
\bm{h}_{S \rightarrow U_t}^{n} =\sum_{k=1}^K \sum_{l=1}^{N(S)} \bm{b}_s^k * \alpha^{nkl}_t \bm{\widetilde{h}}_{s}^n.
\end{equation}

Following Eq. (\ref{lstm}), a BiLSTM is built upon $\bm{h}_{S \rightarrow U_t}^{n}$ to obtain the final utterance representations $\hat{\bm{h}}_{U_t}^n$ which is sent into the decoder.

\subsection{Decoder with Copy Mechanism}
Our decoder is similar with that proposed by \citet{zhang2019editing}.
The representations of current utterance, schema-states, SQL-states and last predicted SQL query are used as the inputs of the decoder.
It should be noticed that all representations of schema column names  and SQL keywords used in \cite{zhang2019editing} are replaced by the schema-state representations and SQL-state representations in our model.
Moreover, when calculating the scores of SQL keywords, we measure the similarity between the decoder hidden states and SQL-states directly without introducing additional module
like the MLP used by \citet{zhang2019editing}.

\section{Experiments}
\subsection{Implementation Details}

All  hidden states in our proposed IST-SQL model had 300 dimensions except the BERT embedder with 768 hidden dimensions.
The head number $K$ was set as 3 heuristically.
When training model parameters, in addition to the average token level cross-entropy loss for all the SQL tokens in an interaction, regularization terms were added to encourage the diversity of the multi-head attentions used in the utterance encoder and schema linking modules. 

The model was implemented using PyTorch \cite{paszke2017automatic}.
We used ADAM optimizer \cite{kingma2014adam} to minimize the loss function.
The BERT embedder was initialized with a pre-trained small uncased BERT model\footnote{https://github.com/google-research/bert}.
All the other parameters were randomly initialized from a uniform distribution between [-0.1, 0.1].
The BERT embedder was fine-tuned with learning rate of $1e-5$ while the other parameters was trained with learning rate of $1e-3$.
An early stop mechanism was used with patient number 10 on the development set.
The best model was selected based on the token-level string matching accuracy on the development set.
The golden last SQL query was fed into our interaction states tracking module at the training stage.
The final IST-SQL model has 448M parameters while the baseline model Edit-Net \cite{zhang2019editing} has 468M parameters.
All code is published to help replicate our results\footnote{https://github.com/runzewang/IST-SQL}.
\subsection{Metrics}
Two metrics, question match accuracy (QM) and interaction match accuracy (IM)  \cite{yu2019sparc}, were used in our evaluation.
QM is the percentage of the queries corresponding to all evaluated questions that are correctly predicted.
While IM is the percentage of interactions with all queries correctly predicted. 
In order to avoid ordering issues, instead of using simple string matching on each predicted query, we followed the method proposed by \citet{yu2018spider}  which decomposed predicted queries into different SQL components such as $SELECT$, $WHERE$, $GROUP BY$, and $ORDER BY$, and computed accuracy for each component using set match separately.

\subsection{Overall Results}
\begin{table}
\small
\centering
\begin{tabular}{lcccc}
\hline
\multirow{2}*{\textbf{Model}} & \multicolumn{2}{c}{\textbf{QM} (\%)}  & \multicolumn{2}{c}{\textbf{IM} (\%)}  \\
& Dev& Test & Dev& Test\\
\hline
SyntxtSQL-con \cite{Yu19SParC} & 15.1 & 14.1 & 2.7 &2.2 \\
CD-Seq2Seq \cite{Yu19SParC}& 13.8 & 13.9 &2.1 &2.6 \\
Edit-Net \cite{zhang2019editing}& 39.9 & 40.8 & 12.3& 13.7\\
\hline
IST-SQL (Ours) & \textbf{44.4} & \textbf{41.8} & \textbf{14.7} &\textbf{15.2} \\
\hline
\end{tabular}
\caption{\label{overall_result_cosql}Results of different models on CoSQL. Due to submission limitations, we only report the results of our model which achieved the best QM performance on the development set.
The results of other models are copied from \cite{yu2019cosql} and \cite{zhang2019editing}.}
\end{table}

\begin{table}
\small
\centering
\begin{tabular}{lcc}
\hline
\textbf{Model} & \textbf{QM} (\%) & \textbf{IM} (\%)\\
\hline
Edit-Net \cite{zhang2019editing} & 45.35 $\pm$ 0.62 & 26.60 $\pm$ 0.70 \\
IST-SQL (Ours)  & \textbf{47.55 $\pm$ 0.80} & \textbf{29.93 $\pm$ 0.93} \\
\hline
\end{tabular}
\caption{\label{sparc_results} Results of different models on SParC development set. Each model was trained four times. The mean and standard deviation of each metric are reported.}
\end{table}
We compared our IST-SQL model  with three baseline models, SyntxtSQL-con \cite{Yu19SParC}, CD-Seq2Seq \cite{Yu19SParC} and Edit-Net \cite{zhang2019editing}.
Because the test set of CoSQL is not publicly  available and there are limitations on the times of online submissions, we only evaluated our model which achieved the best QM performance on the development set.
The results are shown in Table \ref{overall_result_cosql}.
We can see that our IST-SQL model achieved the best development and test performance among all the models on both QM and IM.
Comparing with the state-of-the-art method Edit-Net, our model improved its QM from 39.9\% to 44.4\% and IM from 13.7\% to 15.2\%.
Furthermore, our methods achieves the second place on the leaderboard of CoSQL\footnote{https://yale-lily.github.io/cosql}, while the first place method has not been published yet.

We also compared our model with Edit-Net on the SParC dataset.
Due to time limitation, we haven't received the online test set results till paper submission.
Thus, we only report the QM and IM results on the development set.
Each model was trained four times, 
and the mean and standard deviation are shown in Figure \ref{sparc_results}.
We can find that our model improved the Edit-Net performance from 45.35\% to 47.55\% on QM and from 26.6\% to 29.93\% on IM, while both of the improvements were larger than their standard deviations.
\begin{table*}[t]
\small
\centering
\begin{tabular}{lccccccc}
\hline
\multicolumn{2}{c}{} & \textbf{SELECT} &\textbf{WHERE} & \textbf{GROUP} & \textbf{ORDER}& \textbf{AND/OR} & \textbf{IUEN}\\
\hline
\multicolumn{2}{c}{Clause Number} & 1004 & 573 & 128&165&20&19 \\
\hline
\multirow{2}*{\textbf{F1(\%)}} & Edit-Net & 73.6 & 57.6 & 40.7 & 64.5 & 96.7 & 14.8 \\
& IST-SQL & 73.7& 60.7 & 51.2 & 66.0 & 97.2 & 15.6\\
\hline
\end{tabular}
\caption{\label{sql_clause}The development set F1 (\%) results of two models on different SQL clauses.}
\end{table*}

\subsection{Ablation Study}
\begin{table}[t]
\small
\centering
\begin{tabular}{lcc}
\hline
\textbf{Model} & \textbf{QM} (\%) & \textbf{IM} (\%)\\
\hline
Ours & \textbf{43.05 $\pm$ 0.85} & \textbf{15.33 $\pm$ 0.51} \\
\hline
w/o schema-states & 40.88 $\pm$ 0.83 & 12.60 $\pm$ 0.79 \\
w/o SQL-states & 41.73 $\pm$ 0.89 & 13.70 $\pm$ 1.20 \\
\hline
\end{tabular}
\caption{\label{ablation_study} Results of ablation studies on the two types of interaction states in our model. Each model was trained four times and was evaluated on the development set.}
\end{table}

We further investigated the effects of   proposed schema-states and SQL-states in our model.
Two ablation experiments were performed by removing each type of interaction states from the full model.
When removing schema-states, all schema-state representations used in downstream modules were replaced with the representations of schema column names generated by the BERT embedder.
When removing SQL-states, all SQL-state representations used in downstream modules were replaced with the embedding vectors estimated for all SQL keywords.
For better model comparison, we trained each model four times and calculated the mean and standard deviation of its QM and IM.

The results are shown in Table \ref{ablation_study}.
When removing schema-states from the model, the performance of QM dropped from 43.05\% to 40.88\% while the IM dropped from 15.33\% to 12.6\%.
When removing SQL-states, the performance on OM dropped from 43.5\% to 41.73\% while the IM dropped from 15.33\% to 13.7\%.
Furthermore, all performance degradations 
were larger than their corresponding standard deviations.
These results indicate that both 
schema-states and SQL-states contributed to achieve the overall performance of our IST-SQL model. 

\subsection{Analysis}
In order to better understand the advantages of our IST-SQL model, 
three analysis experiments were further conducted to compare IST-SQL model with the baseline Edit-Net model.
We re-trained the Edit-Net model four times and chose the best one with 41.2\% QM and 13.7\% IM on the development set, which was better than the model shared by its authors.

\paragraph{Difficulty Level}
\begin{table}
\small
\centering
\begin{tabular}{lccccc}
\hline
\multicolumn{2}{c}{} & \textbf{Easy} &\textbf{Medium} & \textbf{Hard} & \textbf{Extra-Hard}\\
\hline
\multicolumn{2}{c}{Utterance Number} & 415 & 320& 162&107\\
\hline
\multirow{2}*{\textbf{QM(\%)}} & Edit-Net &  62.9 & 33.1 & 22.2 & 10.3 \\
& IST-SQL & 66.0 & 36.2 & 27.8 & 10.3 \\
\hline
\end{tabular}
\caption{\label{difficulty level}The development set QM (\%) results of two models on utterances with different difficulty levels.}
\end{table}
We first evaluated the two models, IST-SQL and Edit-Net, on the utterances with different difficulty levels.
Here, the difficulty levels were determined based on the components of target SQL queries, and the decision rules followed previous work \cite{yu2018spider}.
The results are shown in Table \ref{difficulty level}.
We can see that our IST-SQL model  outperformed Edit-Net on  ``Easy", ``Medium" and ``Hard" levels, while obtained the same QM results on ``Extra-Hard" level whose target SQL queries are most complex and usually contain nesting structures.
\paragraph{Turn Index}
\begin{table}
\small
\centering
\begin{tabular}{lcccccc}
\hline
\multicolumn{2}{c}{} & \textbf{1} &\textbf{2} & \textbf{3} & \textbf{4} & \textbf{\textgreater 4}\\
\hline
\multicolumn{2}{c}{Utterance Number} & 292 & 283& 244&114 & 71\\
\hline
\multirow{2}*{\textbf{QM(\%)}} & Edit-Net &  52.1 & 39.6 & 37.3 & 36.0 & 25.4\\
&IST-SQL & 56.2 & 41.0 & 41.0 & 41.2 & 26.8\\
\hline
\end{tabular}
\caption{\label{turn index}The development set QM (\%) results of two models on utterances with different turn indices.}
\end{table}
Then, we evaluated the two models on the utterances with different turn indices.
We split all the interactions in the development set and regrouped them based on their turn indices in interactions for evaluation.
The results are shown in Table \ref{turn index}.
We can find that our model achieved better performance than Edit-Net on utterances with all turn indices.
For our IST-SQL model,  the QM of the utterances at the fourth turn  was  comparable with that of the utterances at the second turn. 
These results demonstrate the ability of  our model on dealing with deep turns by tracking the interaction process with schema-states and SQL-states.

\paragraph{SQL Clause}
As last, we evaluated the development set performance of these two models on different SQL clauses.
The true SQL queries and predicted SQL queries were separated into several clauses with SQL keywords.
Each clause had a value set that was organized following previous study \cite{yu2018spider}.
For each clause, we calculated its F1 score of predicted SQL queries according to true SQL queries.
The results are shown in Table \ref{sql_clause}.
We can find that our model outperformed Edit-Net on all SQL keyword clauses.
The ``AND/OR" and ``SELECT"  clauses were the two easiest ones and the advantages of IST-SQL over Edit-Net were slight on these two clauses.
``GROUP" and ``WHERE" were the two clauses that achieved the largest improvements.
One reason is that these two clauses usually contain more schema column names than other clauses such as ``ORDER"  and ``AND/OR".
Another reason is that these two clauses usually appear at later turns than the clauses such as ``SELECT".
Therefore, these two clauses can benefit most from tracking interaction states in the IST-SQL.


\section{Conclusion}
In this paper, we have proposed an IST-SQL model to deal with the multi-turn text-to-SQL task.
Inspired by the dialogue state tracking component in task-oriented dialogue systems,
this model defines two types of interaction states, schema-states and SQL-states, and their values are updated according to last predicted SQL query at each turn.
Furthermore, a relational graph neural network is employed to calculate schema-state representations while a non-linear layer is adopted to obtain SQL-state representations. 
These state representations are combined with utterance representations for decoding the SQL query at current turn.
Our evaluation results on CoSQL and SParC datasets have demonstrated the effectiveness of the IST-SQL model. 
To improve current definitions and tracking modules of interaction states for conversational text-to-SQL with response generation will be a task of our future work.
\bibliographystyle{aaai21}
\bibliography{aaai21}
\end{document}